# Automated Identification of Tree Species by Bark Texture Classification Using Convolutional Neural Networks

Sahil Faizal[1]

*Student at School of Computer Science and Engineering*

*Vellore Institute of Technology, Chennai*

***Abstract:*** Identification of tree species plays a key role in forestry related tasks like forest conservation, disease diagnosis and plant production. There had been a debate regarding the part of the tree to be used for differentiation, whether it should be leaves, fruits, flowers or bark. Studies have proven that bark is of utmost importance as it will be present despite seasonal variations and provides a characteristic identity to a tree by variations in the structure. In this paper, a deep learning based approach is presented by leveraging the method of computer vision to classify 50 tree species, on the basis of bark texture using the BarkVN-50 dataset. This is the maximum number of trees being considered for bark classification till now. A convolutional neural network(CNN), ResNet101 has been implemented using transfer-learning based technique of fine tuning to maximise the model performance. The model produced an overall accuracy of >94% during the evaluation. The performance validation has been done using K-Fold Cross Validation and by testing on unseen data collected from the Internet, this proved the model's generalisation capability for real-world uses.

***Keywords:*** Image Classification, Data Augmentation, Species Identification, Convolutional Neural Networks, Sampling

## I. INTRODUCTION

Bark is the outermost covering of trees and shrubs plays an essential role in its survival. One of the main functions of bark is to deliver products obtained by photosynthesis to the plant tissues. The protective nature of bark is also notable as it acts as a shield against herbivory and fire and provides insulation from cold weather. Bark usually covers the trunks and branches of trees and is made up of several layers. Its appearance is mainly dependent on cork development. The variation in thickness and texture is often brought in by the environment in which it grows. A lot of commercially and medically important products are derived from the bark of trees.

Researchers in the field of environment, physiology, meteorology etc. have high regard for tree identification. They take into account various parameters like leaf, stem, colour, fruit, flower, bark etc. for the classification. These practices are conducted at high expense and risk by experts having years of related experience and domain knowledge by venturing into the forested areas. Even after these the results are not highly accurate or consistent, as seen in [1] where two experts analysed data collected on Austrian tree species and achieved an accuracy of 56.6% and 77.8% during classification. To solve this problem various works have been done in the related identify tree species on the basis of bark texture classification using techniques of image processing, machine learning and deep learning.

Recent advances in technology have paved the way for deep learning based neural networks to learn and extract features from data to produce accurate results. With the advent of computer vision, machines became capable enough to make predictions from images and videos. However a large amount of data is often required for deep learning algorithms to obtain better performance. Despite being a prime area of importance in agricultural research, tree recognition received limited attention which can be attributed to the lack of related datasets with multiple classes and diversity. From the literature, it is found that databases like the AFF dataset [1] contains only 1,200 images of 11 classes while datasets like BarkNet 1.0 have over 23,000 images of 23 different tree species of Canada.

In this work, I am focusing on classifying the maximum number of tree species we are using with the BarkVN-50[2] dataset consisting of 50 tree species and 5,678 images of 303 x 404 pixels size. For the purpose of multi-class classification a novel architecture of ResNet101 with Back Propagation [3] is proposed, a fine tuning approach has been followed to classify the bark texture and thereby identify the original tree species. The model performance can be accessed by evaluation metrics like accuracy, precision, recall and f1-score.

This paper is organised into five sections. In Section II, a literature survey on related works is done to understand the methods used and the results obtained. This is followed by Sections III and IV consisting of the details regarding the proposed techniques, data and tools used for the purpose of bark texture classification. Finally, Section V contains the detailed report of the experiments performed and the results obtained while Section VI concludes the work by noting down the key achievements, limitations and the scope for improvement.

## II. RELATED WORKS

Different types of approaches are adopted for the classification of texture; some include using traditional machine learning based techniques along with image processing techniques for texture feature analysis. However not much work has been done using Convolutional Neural Networks(CNNs)[4], which have now become the pioneers for feature extraction from images. Some of them include filtering methods like Fourier and Gabor Filters , morphological filters , wavelet based [5,6] etc. A texture descriptor, statistical macro binary pattern (SMBP) [7] is used for extracting statistical information and performing neighbourhood sampling helped them achieve an accuracy 71%.

In [8] a technique of feature extraction is done by combining textural features through GLCM(Grey level co-occurrence matrix), which is commonly used for second order textural analysis and fractal dimension features then using the feature vectors the bark images are classified into their respective classes using Artificial Neural Networks(ANNs). Automated identification of plant species have also grown over the years where supervised machine learning [9] and deep learning techniques based on transfer learning[10] are put to use.

M.Carpentier et al.[11] made a custom dataset, BarkNet-1.0 with 23,000 images covering 23 different tree species of Canada. This data is then used to train a ResNet architecture using pre-trained weights from the ImageNet dataset(made up of over 1 million images of around 10,000 classes of objects) for weight initialisation. From experimentation it was seen that the model accuracy varied between 93.88% and 97.81%. Terrestrial Laser Scanning (TLS) method is used in [12] to obtain 3D point clouds for identifying individual tree species from mixed plantations. To calculate roughness measures and shape attributes, the three-dimensional geometric texture of the bark is analysed and then fed as input to the Random Forest classifier for tree species classification. Through this method of 3D based technology accuracy ranges between 83% and 100% is obtained for different tree species.

Application of class activation mapping is seen in [13], where two convolutional neural networks with different architectures are proposed to classify 42 species of trees and achieve an accuracy above 90%. Class activation mapping(CAM) enables modifications in some parameters of CNN architecture and highlights the influential regions used for the prediction purpose. Textural features extracted through multispectral spiral wide-sense Markov model, which benefits from full descriptive colour and rotational invariance are applied for tree bark identification in [14]. The performance evaluation has been done using datasets like the BarkNet, BarkText, Trunk12 and AFF.

A portable tree identification system, Deep BarkID[15] which is deployed on a smartphone with no requirement for database connection and internet access has performed well in classifying 10 tree species in Indiana, USA. The authors have also made a dataset, the Indiana Bark Dataset with thousands for images and is available for public use. A new method based on the application of Gabor filters has been used with different orientations and scales for feature extraction is seen in [16]. Further the feature vectors are fed as input to RBPNN and SVM classifiers for bark classification. The recognition rate by using distinct colour spaces like RGB, HSV, Grey scale etc. has also been documented in this work.

## III. PROPOSED METHODOLOGY

In this work tree species are identified by classifying the bark texture of trees, which is considered important since it remains the same and is not susceptible to seasonal changes. Classifying into multiple classes has always been a challenge faced by researchers, but with the advent of technology techniques like deep learning gained popularity since we have more computational power and data available for training huge networks.

It is an undeniable fact that the power of deep learning based networks to learn increases with the depth of the network and by the amount of data. Here in this research we have a dataset containing 50 classes of trees and 5780 images; but this quantity of data ie; images we have per class is in range 60 - 220, which is not enough to train a Convolutional Neural Network(CNN) based classifier from scratch. The problem is further intensified by the number of classes into which the images should be classified. To solve this problem the technique of Transfer Learning is introduced, in which the feature extracting experience gained by pre-built networks after training on various high volume datasets like ImageNet, COCO etc. can be utilised as per use cases. In short a model created for one specific task is reused for another.

Transfer learning based approach of Fine-Tuning is used, which allows chances in the architecture apart from reusing the pre-built neural network. The method of fine tuning offers an advantage here as the fully connected layers or the outer layers of the deep neural network architecture can be tuned or replaced by the specific use case here, where the image feature extract part by convolutional as well as other related layers remains intact without any changes. This way the feature vectors extracted by the first part through transfer learning can be used by the classification layers for classifying the bark texture into 50 classes. Before training the network is instantiated with the pre-trained weights obtained during the training using ImageNet dataset, which contains over 1 million images and more than 10,000 classes.

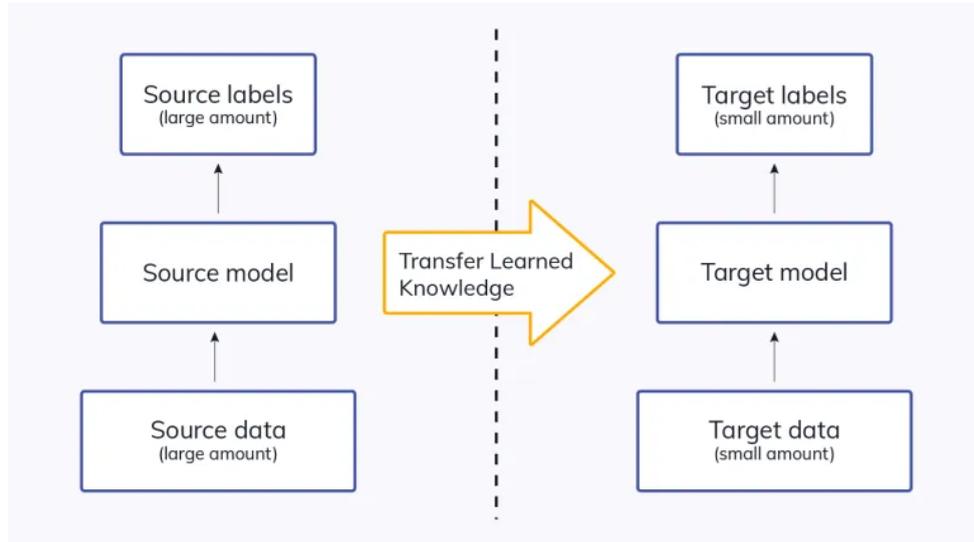

Fig 1. Idea behind transfer learning

The challenge posed by the class imbalance problem must also be addressed because even if the model performs well during the training phase, there is a significant probability that the model might show some bias towards those classes with more samples. This happens because of the fact that the model has seen the common data more compared to the ones having a small proportion of data. In order to solve this the technique of Data Augmentation has been performed to, where the proportion of images per class is increased by artificially generating more samples by the application of image processing techniques. Here the Python package of Augmentor is used to generate synthetic images using the same data. The count is increased to a limit of 110 images per class as the median and mean values of the image count per class of the original dataset was in a similar range. This task thus involved random oversampling and undersampling.

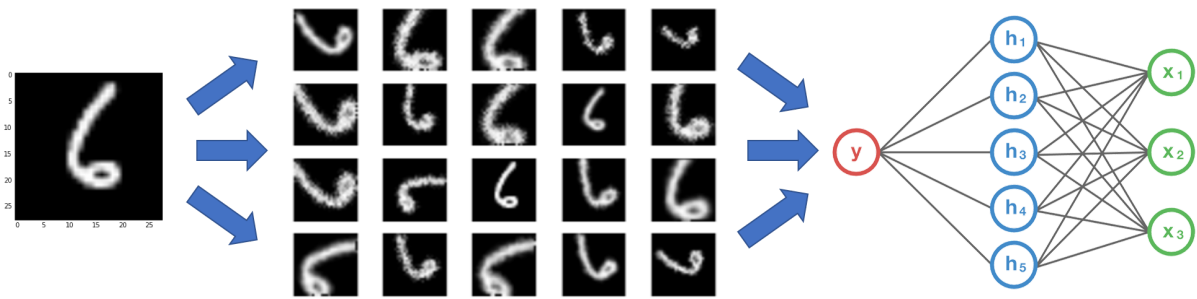

Fig 2. An example of data augmentation

The classes containing more images are considered as majority classes and ones with minimum images are termed as minority classes. Through the approach adopted here, the required amount of data is randomly selected from the majority class while minority class data gets augmented. Hence the majority class undergoes undersampling while the minority class undergoes oversampling. This will also preserve the length of the original dataset as there will be 5500 images instead of 5678, which is not a drastic difference. These images are then resized to 160 x 160 pixels to reduce the complexity as well as the computational power required for

training. Since three channels(Red(R), Green(G) and Blue(B)) are used the input array dimension becomes :-
h * w * 3, where h is the image height and w is the image width.

The computational power required for training the model in this work is still significant as large deep learning based networks have higher processing and memory(RAM) needs which traditional PC machines fail to offer. A GPU-accelerated training using NVIDIA GPUs is always beneficial as it reduces time taken. Users can either prefer a GPU based computer or a cloud platform for the training. Various cloud based platforms are available for training like Google Cloud Platform(GCP), Amazon AWS, Microsoft Azure, IBM Cloud etc. But the most commonly used platform for research purposes is Google Colaboratory(Google Colab).

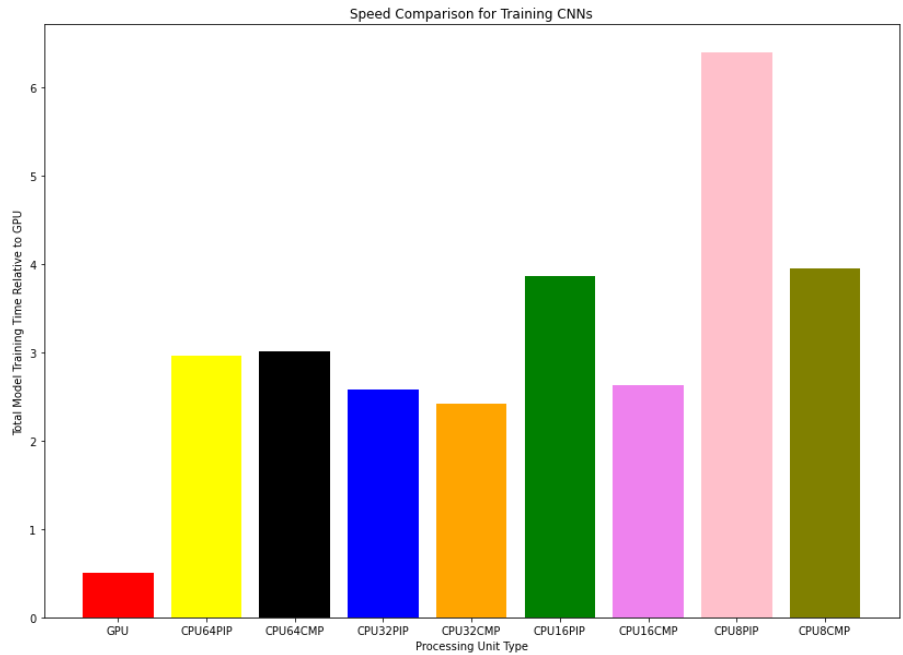

Figure 3. Comparison of difference in training speeds

## IV. PROJECT WORKFLOW

The proposed work is divided into four parts - Data Collection, Resampling, Data Pre-Processing and Model Building. Modularisation has been done so that project management can be effective and help in reducing the time taken for debugging the errors in the code base.

*A. Data Collection:-*
The image data, BarkVN-50 used in this work has been collected by Vinh Truong Hoang of the Ho Chi Minh City Open University for the research purpose. And has been provided for open source access [2]. A total of 5,768 images of 50 tree categories are present in this dataset. The images represent the bark textures of the trees and are of dimensions 303 x 404 pixels. The dataset has been specifically made for the purpose of image classification.

*B. Re-sampling:-*
After exploring the contents of the dataset it was noticed that there existed a significant imbalance in the dataset, which could have affected the model performance by introducing a bias towards classes with more samples.
It was not feasible to either perform only one among the traditional techniques of upsampling or downsampling since the motive was to preserve the class ratio and the original size of the dataset. Hence a mixture of both has been applied to produce a sample strength of 5500 images divided into 50 classes ie; 110 images per class.
Image processing operations like zoom, flip top-bottom, random distortion and random brightness change has been done to synthetically create new data by setting a probability parameter for each action to take place.

*C. Data Pre-Processing:-*
The re-sampled images are further pre-processed by resizing them into 160 x 160 pixels RGB images for the ease of computation. All the three channels R, G and B of the colour image are used without converting to grayscale as colour can also act as a differential factor for some species. The pixel intensity values are then scaled down to be in range 0 and 1 for faster processing and to reduce any bias. The data is then stored in numpy arrays for easy access and operations.

*D. Model Building:-*
A transfer learning based approach, fine tuning is followed in the model building process. ResNet-101, having 101 layers, is used as the base model for extracting distinctive features from the images. The fully connected layers are custom built for the purpose of classification. A total of 4 fully connected layers are there, in which the first two have dropout layers in between which cut off 45% of the connections in between them. The dropout layers are used to control the model overfitting. There are 512 neurons each in the first and second layers, while the penultimate layer has 256 neurons whereas the final output layer has 50 neurons representing the classes. Each of the first three layers have relu activation to bring in non-linearity in the network. For aiding in the process of backpropagation, Adam optimizer along with categorical loss entropy loss function has been used. The final layer has softmax activation function, which gives a probabilistic output of the probable class to which the image belongs to.

*E. Model Evaluation:-*

After the process of training the model has been evaluated through metrics like Accuracy, Precision, Recall and F1-score. K-Fold Cross Validation has been performed here to validate the model performance and to assess the models capability to generalise; in the process the entire dataset is split into k folds are then fed as input to the neural network the respective performance scores after each fold is stored for evaluation purpose. A holistic evaluation has been done using classification reports, and from the average performance scores obtained from the cross validation.

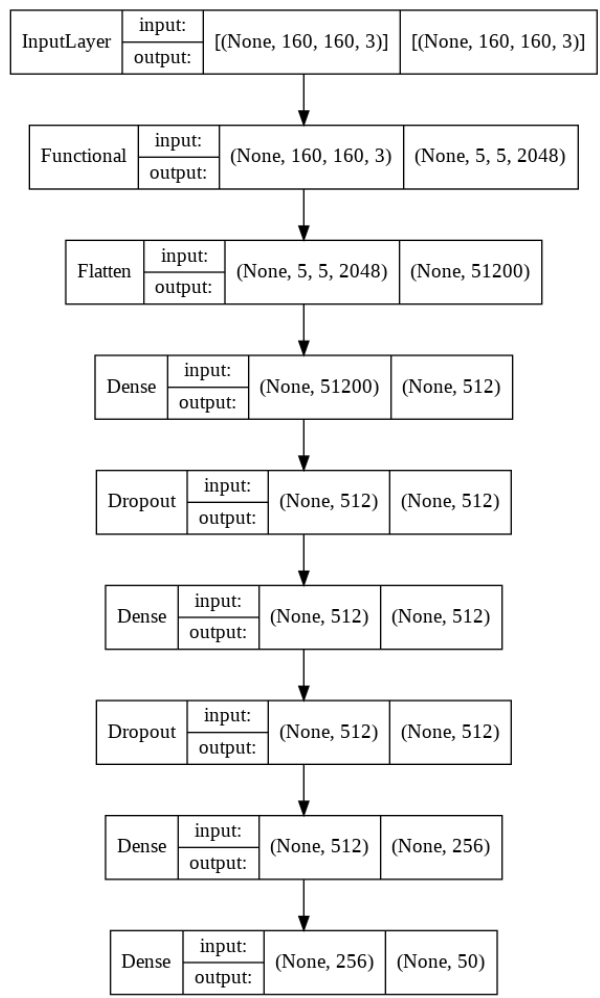

Fig 4. Model Plot

```
Model: "sequential"
Layer (type)                 Output Shape              Param #
=================================================================
resnet101v2 (Functional)     (None, 5, 5, 2048)        42626560
flatten (Flatten)            (None, 51200)             0
dense (Dense)                (None, 512)               26214912
dropout (Dropout)            (None, 512)               0
dense_1 (Dense)              (None, 512)               262656
dropout_1 (Dropout)          (None, 512)               0
dense_2 (Dense)              (None, 256)               131328
dense_3 (Dense)              (None, 50)                12850
=================================================================
Total params: 69,248,306
Trainable params: 69,150,642
Non-trainable params: 97,664
```

Fig 6. Model Summary

## V. EXPERIMENTATION AND RESULTS

The images used in this experiment are of dimension 160 x 160 pixels with depth of 24 bits/pixel. The image data is stored in the numpy array after stages of re-sampling and pre-processing are then split into training and testing sets in the ratio 80:20 respectively.

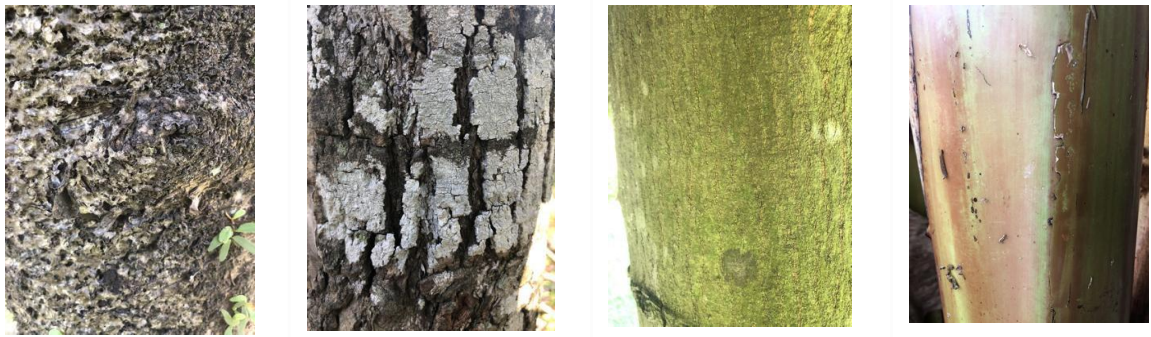

Fig 7. Four kinds of bark images

For feature extraction as mentioned in Section IV, the model is relying heavily on the transfer learning based approach where the convolution base has prior experience in creating feature vectors from images. The model configuration has been clearly outlined in Fig 5. For model training, Google Colaboratory has been used which provides on demand cloud resources for research tasks. The platform has allocated a 16 GB memory version of NVIDIA Tesla P-100 GPU along with 25 GB RAM and 170 GB of storage. The provision of GPU will enable distributed processing of data and ensure faster training speeds for the experimental purpose.

The best model architecture for this work has been chosen after experimenting with prominent CNN based networks like VGG 19, ResNet-50, ResNet-101, InceptionV3 and MobileNetV4, in which ResNet-101 was better in terms of overall performance measured in terms of accuracy. The learning rate of the Adam optimizer has been set to 0.0001 initially. The model is instantiated using the pre-trained weights from ImageNet classification. These weights are subjected to change during the training by weight updation through the process of backpropagation. The model is trained for different epochs in range 20 - 200 and the respective measures of accuracy, loss are stored. An epoch value between 30 and 40 is seen as optimal as there is no significant change in performance afterwards.Through analysis epoch value of 32 is chosen and obtained an overall accuracy of 99.40% and 94.3% in the phases of training and testing respectively.

TABLE I
Testing Phase Results

| Accuracy | Precision | Recall | F1-Score |
|---|---|---|---|
| 94.36% | 94.66% | 94.37% | 94.33% |

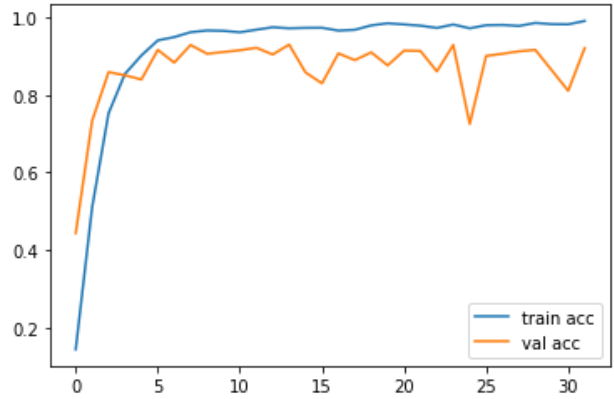
Fig 8. Accuracy vs Epoch Curve

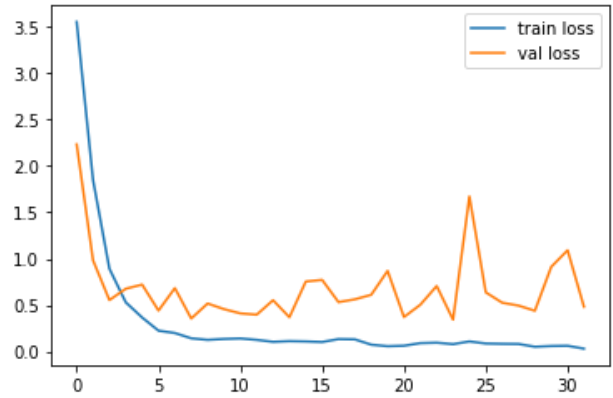
Fig 9. Loss vs Epoch Curve

Graphs showcasing the accuracy and loss throughout the training process have been plotted to understand how the loss has gone down and the accuracy got improved over time. In Fig 8,9 the plots are made in such a way that the epoch count is represented by the X-axis while the Y-axis represents accuracy and loss respectively.

|  | precision | recall | f1-score | support |
|---|---|---|---|---|
| Wrightia religiosa | 0.95 | 1.00 | 0.97 | 18 |
| Wrightia | 1.00 | 0.96 | 0.98 | 24 |
| Melaleuca | 1.00 | 0.80 | 0.89 | 15 |
| Prunus salicina | 1.00 | 0.95 | 0.98 | 22 |
| Spondias mombin L | 0.90 | 1.00 | 0.95 | 19 |
| Cocos nucifera | 0.96 | 1.00 | 0.98 | 26 |
| Pterocarpus macrocarpus | 0.95 | 0.90 | 0.93 | 21 |
| Melia azedarach | 1.00 | 1.00 | 1.00 | 19 |
| Dalbergia oliveri | 0.85 | 1.00 | 0.92 | 23 |
| Prunnus | 1.00 | 0.94 | 0.97 | 18 |
| Khaya senegalensis | 0.95 | 1.00 | 0.98 | 21 |
| Acacia | 0.94 | 0.94 | 0.94 | 18 |
| Tectona grandis | 0.91 | 0.78 | 0.84 | 27 |
| Gmelina arborea Roxb | 1.00 | 0.94 | 0.97 | 17 |
| Annona squamosa | 1.00 | 0.96 | 0.98 | 23 |
| Cananga odorata | 0.95 | 1.00 | 0.98 | 20 |
| Artocarpus heterophyllus | 0.94 | 0.94 | 0.94 | 18 |
| Musa | 1.00 | 1.00 | 1.00 | 17 |
| Hopea | 0.89 | 0.85 | 0.87 | 20 |
| Eucalyptus | 0.96 | 0.92 | 0.94 | 25 |
| Barringtonia acutangula | 0.95 | 1.00 | 0.98 | 20 |
| Mangifera | 0.88 | 1.00 | 0.93 | 21 |
| Ficus racemosa | 0.91 | 0.91 | 0.91 | 23 |
| Persea | 1.00 | 1.00 | 1.00 | 28 |
| Ficus microcarpa | 0.96 | 1.00 | 0.98 | 24 |
| Artocarpus altilis | 1.00 | 1.00 | 1.00 | 23 |
| Polyalthia longifolia | 1.00 | 0.91 | 0.95 | 23 |
| Carica papaya | 0.86 | 0.95 | 0.90 | 19 |
| Anacardium occidentale | 0.95 | 1.00 | 0.97 | 19 |
| Psidium guajava | 0.96 | 1.00 | 0.98 | 22 |
| Citrus aurantiifolia | 1.00 | 1.00 | 1.00 | 24 |
| Magnolia alba | 0.79 | 0.96 | 0.87 | 24 |
| Cedrus | 0.95 | 1.00 | 0.98 | 21 |
| Adenium species | 1.00 | 1.00 | 1.00 | 21 |
| Adenanthera microsperma | 1.00 | 0.76 | 0.86 | 21 |
| Chrysophyllum cainino | 1.00 | 0.95 | 0.98 | 22 |
| Lagerstroemia speciosa | 0.86 | 0.89 | 0.87 | 27 |
| Terminalia catappa | 0.78 | 0.82 | 0.80 | 22 |
| Dipterocarpus alatus | 0.96 | 0.96 | 0.96 | 25 |
| Khaya senegalensis A.Juss | 1.00 | 1.00 | 1.00 | 25 |
| Casuarina equisetifolia | 0.96 | 0.96 | 0.96 | 23 |
| Tamarindus indica | 0.83 | 1.00 | 0.91 | 15 |
| Citrus grandis | 1.00 | 0.84 | 0.91 | 31 |
| Erythrina fusca | 0.92 | 0.96 | 0.94 | 24 |
| Senna siamea | 0.81 | 0.72 | 0.76 | 18 |
| Delonix regia | 0.82 | 0.88 | 0.85 | 26 |
| Syzygium nervosum | 1.00 | 1.00 | 1.00 | 24 |
| Veitchia merrilli | 0.95 | 1.00 | 0.98 | 21 |
| Nephelium lappaceum | 1.00 | 0.88 | 0.93 | 32 |
| Hevea brasiliensis | 1.00 | 1.00 | 1.00 | 21 |
| accuracy |  |  | 0.94 | 1100 |
| macro avg | 0.95 | 0.94 | 0.94 | 1100 |
| weighted avg | 0.95 | 0.94 | 0.94 | 1100 |

Fig 10. Classification Report

Further analysis of the respective evaluation metrics per class can be obtained from the classification report which depicts the identification and generalisation capability of the model in general. From Fig 10, an overall idea of the model's performance on unseen data can be seen as the model is able to classify the different image instances into their respective groups precisely. Here, support represents the count of actual occurrences of the particular class in the test set.

The results obtained from K-Fold Cross Validation are also satisfactory, k value of 5 has been set to split the dataset into 5 distinct folds and are noted down in Table II.

TABLE II
K-Fold Cross Validation Performance Report

|  | Fold - 1 | Fold - 2 | Fold - 3 | Fold - 4 | Fold - 5 | Average |
|---|---|---|---|---|---|---|
| Testing Accuracy | 94.01% | 94.30% | 94.09% | 94.11% | 94.08% | 94.118% |
| Testing Precision | 94.65% | 94.51% | 94.63% | 94.59% | 94.60% | 94.596% |
| Testing Recall | 94.49% | 94.23% | 94.07% | 94.11% | 94.20% | 94.22% |

## VI. CONCLUSION

In this paper, a deep learning based model capable of classifying 50 different species of trees from its bark texture has been presented. Here the feature extraction is done by convolutional neural networks, which has state of the art architectures for obtaining feature vectors from image data of any dimension and quality. Through the evaluation phase the generalisation ability of the model and its capability to precisely classify has been validated. Moreover, different architectures based on CNN have been explored and experimented with to choose the best performing one ie; ResNet101 here. Further the accuracy, precision and recall measures obtained in this work >90% is higher in comparison to other works since the highest number of classes is considered here. There is still scope of improvement as more data can turn useful for training purposes. In addition to that, data can be collected from different lighting and weather conditions so that the models get ready for real-world deployment. Also different image processing techniques like using filtering techniques, wavelet transforms etc. for extracting more textural features from the bark images can be tried out before feeding as input to the CNN networks. Usage of advanced techniques like Attention-Based mechanisms, CNN-RNN combination and LSTMs for image classification can also be explored.